\documentclass[9pt,conference]{IEEEtran}
\IEEEoverridecommandlockouts

\usepackage{cite}
\usepackage{amsmath,amssymb,amsfonts}
\usepackage{algorithmic}
\usepackage{graphicx}
\usepackage{textcomp}
\usepackage{hyperref}
\usepackage[table]{xcolor}
\usepackage[ruled]{algorithm2e}
\usepackage{booktabs,cite}
\usepackage[super]{nth}
\usepackage{booktabs}
\usepackage{multirow}
\usepackage{multicol}
\usepackage{bbding}
\usepackage{subcaption}    
\usepackage{url}
\usepackage{balance}

\newcommand{\DefinedAI}{ContactCenter}

\definecolor{Gray}{gray}{0.9}

\def\BibTeX{{\rm B\kern-.05em{\sc i\kern-.025em b}\kern-.08em
    T\kern-.1667em\lower.7ex\hbox{E}\kern-.125emX}}
\begin{document}

\title{Performance Evaluation of SLAM-ASR: The Good, the Bad, the Ugly, and the Way Forward}


\author{
    \parbox{\linewidth}{%
        Shashi Kumar$^{\small{\clubsuit},1,2}$, 
        Iuliia Thorbecke$^{\small{\clubsuit},1,3}$, Sergio Burdisso$^{\small{\clubsuit},1}$,  
        Esaú Villatoro-Tello$^{\small{\clubsuit},1}$, Manjunath K E$^{4}$, Kadri Hacioğlu$^{4}$ 
        \centering Pradeep Rangappa$^{1}$, Petr Motlicek$^{1,5}$, Aravind Ganapathiraju$^{4}$ and Andreas Stolcke$^{4}$ \\[2ex]
        \small $^{1}$\textit{Idiap Research Institute, Switzerland};
        \small $^{2}$\textit{EPFL, Switzerland}; 
        \small $^{3}$\textit{University of Zurich, Switzerland};\\
        \small $^{4}$\textit{Uniphore, U.S.A. \& India}; 
        \small $^{5}$\textit{Brno University of Technology, Czech Republic}
    }
}

\maketitle

\begin{abstract}
Recent research has demonstrated that training a linear connector between speech foundation encoders and large language models (LLMs) enables this architecture to achieve strong ASR capabilities. 
Despite the impressive results, it remains unclear whether these simple approaches are robust enough across different scenarios and speech conditions, such as domain shifts and speech perturbations.
In this paper, we address these questions by conducting various ablation experiments using a recent and widely adopted approach called SLAM-ASR.  We present novel empirical findings that offer insights on how to effectively utilize the SLAM-ASR architecture across a wide range of settings. Our main findings indicate that SLAM-ASR exhibits poor performance in cross-domain evaluation settings. Additionally, speech perturbations on in-domain data, such as changes in speech rate or additive noise, can significantly degrade performance.
Our findings offer critical insights for fine-tuning and configuring robust LLM-based ASR models, tailored to different data characteristics and computational resources.

\end{abstract}

\begin{IEEEkeywords}
ASR, LLMs, embeddings, speech-to-text alignment, foundation models.
\end{IEEEkeywords}

\renewcommand*{\thefootnote}{\fnsymbol{footnote}}
\section{Introduction}\footnotetext{$^\clubsuit$ Corresponding authors: \{\textit{shashi.kumar, iuliia.nigmatulina, sergio.burdisso, esau.villatoro\}@idiap.ch}}
\renewcommand*{\thefootnote}{\arabic{footnote}}

Enabling large language models (LLMs) to ``comprehend'' non-textual modalities has received substantial attention recently. For instance, in \cite{zhu2023minigpt}  the authors trained a projection layer to align the outputs of a visual encoder with an LLM. In the context of automatic speech recognition (ASR), some early methods utilize a cascaded approach, where speech is first transcribed using an automated ASR system, followed by processing the resulting text with an LLM to enhance the transcription accuracy \cite{huang2024multilingual, li2023prompting, ma2023can, yang2023generative}, or extracting further knowledge from automatic transcripts for downstream tasks~\cite{9746563, 10022718}. However, cascaded approaches have several limitations, including error propagation and lack of valuable paralinguistic information conveyed by acoustics, such as prosody and speaker characteristics. 

Recently, systems that integrate robust speech encoders with instruction-tuned LLMs through a connector/projector layer have been proposed as end-to-end ASR solutions \cite{fathullah2024prompting, das2024speechverse, tangsalmonn, wu2023decoder, ma2024embarrassingly}, henceforth referred to as LLM-based ASR systems.
Intuitively, the main task of the connector/projector is to learn how to transform acoustic embeddings from speech encoders into speech representations (tokens) that are meaningful within the LLM's embedding space. These representations are then combined with text instructions (\textit{i.e.}, prompts) and fed into an LLM to generate various predictions, such as transcription, emotion classification, language identification, and named entity recognition.
Three immediate advantages of such architectures are: (a) computational efficiency, as the entire system can be adapted to new tasks by adopting parameter-efficient approaches, and/or by training the projector layer; (b) efficient use of the vast corpora used in the pretraining of foundation models; and (c)  enhanced generalization capability, as LLMs can leverage prompts for zero-shot or in-context learning to handle unseen tasks effectively \cite{das2024speechverse}.

In \cite{fathullah2024prompting}, the authors proposed attaching an audio encoder (comprising 36 Conformer \cite{gulati2020conformer} layers) to an LLM (Llama-7B \cite{touvron2023llama}) to perform the ASR task. Embeddings generated by the audio encoder are stacked and projected onto the LLM's input embedding space, which is further trained using the low-rank adaptation (LoRA) approach \cite{hu2022lora}. Similarly, in SpeechVerse \cite{das2024speechverse} and SpeechLLM \cite{SpeechLLM_2024}, the authors describe robust multitask training and curriculum learning frameworks that combine pretrained speech and text foundation models via a small set of learnable parameters. Contrary to \cite{fathullah2024prompting}, these approaches applied a 1-D convolution module that operates over the audio feature sequence to ensure compatibility with the LLM. In the end, only the convolutional downsampling module and the LoRA adapter are trained.
\begin{figure}[t]
    \centering
    \includegraphics[width=0.7\linewidth]{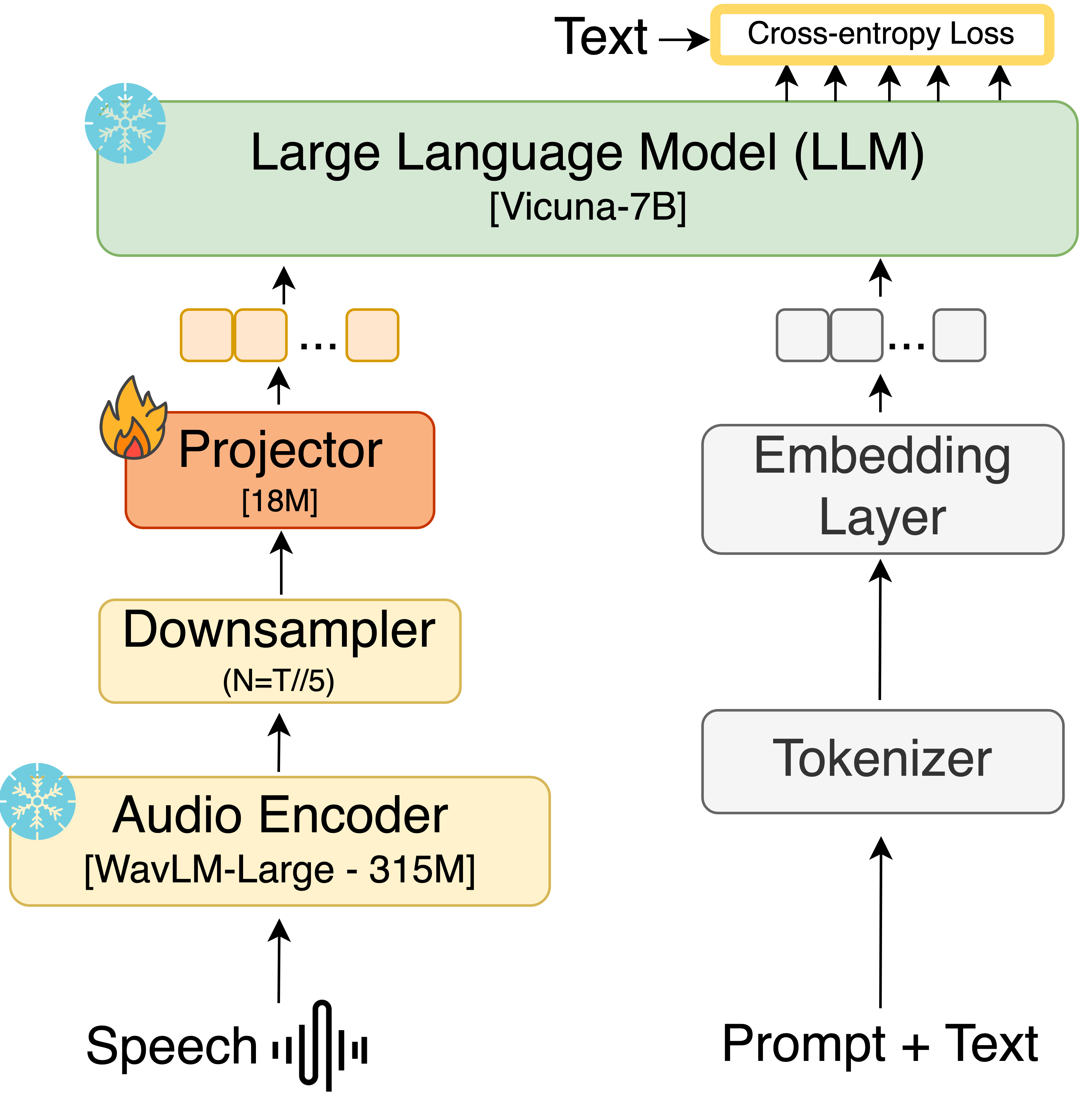}
    \caption{SLAM-ASR pipeline. The selected models and the number of parameters for the performed experiments appear between brackets.}
    \label{fig:SLAM-reloaded}
    \vspace{-6mm}
\end{figure}

Recently, the SLAM-ASR architecture was introduced as ``an embarrassingly simple approach for large language models with robust ASR capabilities'' \cite{ma2024embarrassingly}. 
The authors argue that elaborate neural architecture designs are unnecessary, showing that a simple composition of off-the-shelf speech encoders and LLMs—using only a simple trainable projector to connect them—is sufficient (Figure \ref{fig:SLAM-reloaded}).

Despite the impressive results reported for SLAM-ASR~\cite{ma2024embarrassingly}, it remains unclear whether this novel, yet simple solution, is ``the way to go'' for LLM-based ASR systems.
This paper aims to address this question and performs three ablation experiments on the SLAM-ASR architecture to evaluate how strong its claimed ASR capabilities really are.
We study and analyze the performance of SLAM-ASR on three well-known benchmark datasets and one private dataset, as well as its performance under extreme, yet plausible conditions normally addressed by traditional ASR models.
This allows insights into the structure and organization of the knowledge represented within the SLAM-ASR architecture, providing transparency and interpretability of the network’s behavior.

Overall, our work makes three main contributions that we hope will support the methodological decisions of future researchers working on LLM-based ASR in the SLAM-ASR paradigm: \textit{(i)} we show empirically that SLAM-ASR has a big dependence on the data used for training the projector (\textit{i.e.}, overfitting), resulting in a model that lacks robustness when training and test data are mismatched; \textit{(ii)} we show how sensitive the SLAM-ASR architecture is to temporal and noise perturbations, unlike its ASR counterpart, and \textit{(iii)} we performed an exhaustive analysis, both quantitative and qualitative, of what type of alignments the projector layer is learning and, we show how such an alignment can be improved to give better ASR performance.

\section{Methods}
\label{secc:methods}
\vspace{-2mm}
\subsection{SLAM-ASR}
\label{subsec:slam-asr}
The SLAM-ASR architecture consists of a speech encoder, followed by a fixed downsampler, then a trainable projector, with the final network being an instruction-tuned LLM accepting text embedding (prompt) and audio representations coming from the projector (Figure~\ref{fig:SLAM-reloaded}).
One advantage of this implementation is its flexibility, making it possible to use different existing pretrained speech encoders, as well as a variety of LLMs.
According to the original work~\cite{ma2024embarrassingly} and confirmed by our experiments, the combination of \textit{WavLM-large} \cite{chen2022wavlm} as a speech encoder and \textit{Vicuna-7B} as an LLM achieves the best performance.
Thus, in the experiments reported here, we always use the same \textit{WavLM-large+Vicuna-7B} combination as the original work.
The key element of the SLAM-ASR architecture is that the only trained element is the \textit{projector} \footnote{Although referred to as a `linear projector' in the original work, it employs a non-linear activation function (ReLU) as shown in Eq.~\ref{eq:projector}. In this work, we refer to it simply as \textit{projector} to avoid confusion.},
a single hidden layer followed by a ReLU activation and a regression layer:
\begin{equation}
\label{eq:projector}
    \mathbf{E}_i = Linear(ReLU(Linear(\mathbf{Z}_i)))
\end{equation}where $\mathbf{Z}_i$ is the $i$th downsampled audio feature, consisting of the concatenation of $k$ consecutive frames (output by WavLM) in the feature dimension, and $\mathbf{E}_i$ has the same dimension as the LLM input embedding. We refer to $\mathbf{E}_i$ as the \textit{speech token embedding} of the $i$th audio feature.
Finally, an input audio consisting of the $n$ downsampled audio features $\mathbf{Z}_0\cdots \mathbf{Z}_n$ is given to the instruction-tuned LLM using the following prompt:\footnote{In the original SLAM-ASR paper, speech embeddings are placed after \texttt{"USER:"}, but in the released code, they are prepended to the input; we adopt the latter prompt format .}


\begin{description}
    \item $\mathbf{E}_0\cdots \mathbf{E}_n$\texttt{<s>USER: Transcribe speech to text.}
    \item \texttt{ASSISTANT:\textit{\{transcript\}</s>}}
\end{description}
where \textit{\texttt{\{transcript\}}} is the text corresponding to the audio, which is either provided during training or forced to be generated by the LLM at inference time.

\subsection{Baseline model}
\label{subsecc:baseline}

For the sake of a fair comparison, we will use the \textit{WavLM-large}~\cite{chen2022wavlm} based model as our baseline ASR system since, as in the original work, is also used as the speech encoder in the SLAM-ASR model.
More precisely, the baseline simply consists of adding a linear layer on top of the \textit{WavLM-large}, trained using Connectionist Temporal Classification (CTC) \cite{graves2006connectionist} loss, as a typical End-to-End (E2E) ASR model.
We refer to this model simply as ``ASR''.
\begin{table}[t!]
    \centering
    \caption{Cross-domain performance comparison between the baseline (ASR) and SLAM-ASR in terms of \textit{word error rate} (WER). $\infty$ denotes WER values greater than 100.}
    \begin{tabular}{l@{~~}c@{~~}c@{~~}c@{~~}c}
        \toprule
        \textbf{Training}& \multicolumn{4}{c}{\textbf{Evaluation Datasets} (WER ($\downarrow$))}\\
        \cmidrule(lr){2-5}
         \textbf{Data} & \textit{LibriSpeech} & \textit{CallHome} & \textit{\DefinedAI} & \textit{CommonVoice}\\
         \midrule      
         \rowcolor{Gray} \multicolumn{5}{l}{\textbf{\quad \textit{ASR}}} \\
         \midrule 
         LibriSpeech & \textbf{2.6} & 38.5 & 31.8 & 26.2 \\
         CallHome    & 15.9& \textbf{25.9} & 30.5 & 48.9\\
         \DefinedAI   & 17.1& 37.2 & \textbf{17.0} & 44.1\\
         \midrule      
         \rowcolor{Gray} \multicolumn{5}{l}{\textbf{\quad \textit{SLAM-ASR}}} \\
         \midrule 
         LibriSpeech & \textbf{2.6} & $\infty$ & 69.7 & 51.5 \\
         CallHome    & 61.8 & \textbf{35.5}   & 44.5 & $\infty$\\
         \DefinedAI   & 60.4 &  $\infty$    & \textbf{13.8} & 67.7 \\
         \bottomrule
    \end{tabular}
    \label{tab:cross_domain_results}
    \vspace{-4mm}
\end{table}
\section{Experimental Setup}
\vspace{-2mm}
\subsection{Datasets}
\label{subsecc:dataset}
We selected three well-known benchmark datasets---LibriSpeech, CallHome, and CommonVoice---due to their distinct characteristics and the unique challenges they present.
Additionally, we experimented with a private dataset, \DefinedAI, composed of multidomain contact center conversations.

\noindent \textit{LibriSpeech} consists of read English speech derived from audiobooks with 1000 hours of speech sampled at 16 kHz \cite{panayotov2015librispeech}. We trained our models using the 960h training partition, while all the evaluations were performed in the LibriSpeech \textit{test-clean} partition. 

\noindent \textit{CallHome English} (LDC97S42) contains spontaneous telephone conversations between multiple speakers, comprising 12.5h of transcribed training speech and 1.5h of test data. This dataset poses challenges due to its conversational nature, known to be difficult for ASR, with a large number of short segments.

\noindent \textit{CommonVoice} comprises several thousand hours of crowdsourced audio in more than 100 languages \cite{ardila2020common}, featuring significant variability in speakers, accents, speaking styles, and background noise, among others. We used the English test partition from CommonVoice-v11, containing 27h, but only as test set, to evaluate robustness.

\noindent \textit{\DefinedAI}
contains 48h of training and 6h of test stereo-audio/transcript data from contact center conversations between agents and customers. We upsampled audio from 8\,kHz to 16\,kHz to align with the SLAM-ASR model's requirements.

\begin{figure*}[h]
\centering
    \begin{subfigure}{.3\textwidth}
        \centering
        \includegraphics[width=.95\linewidth]{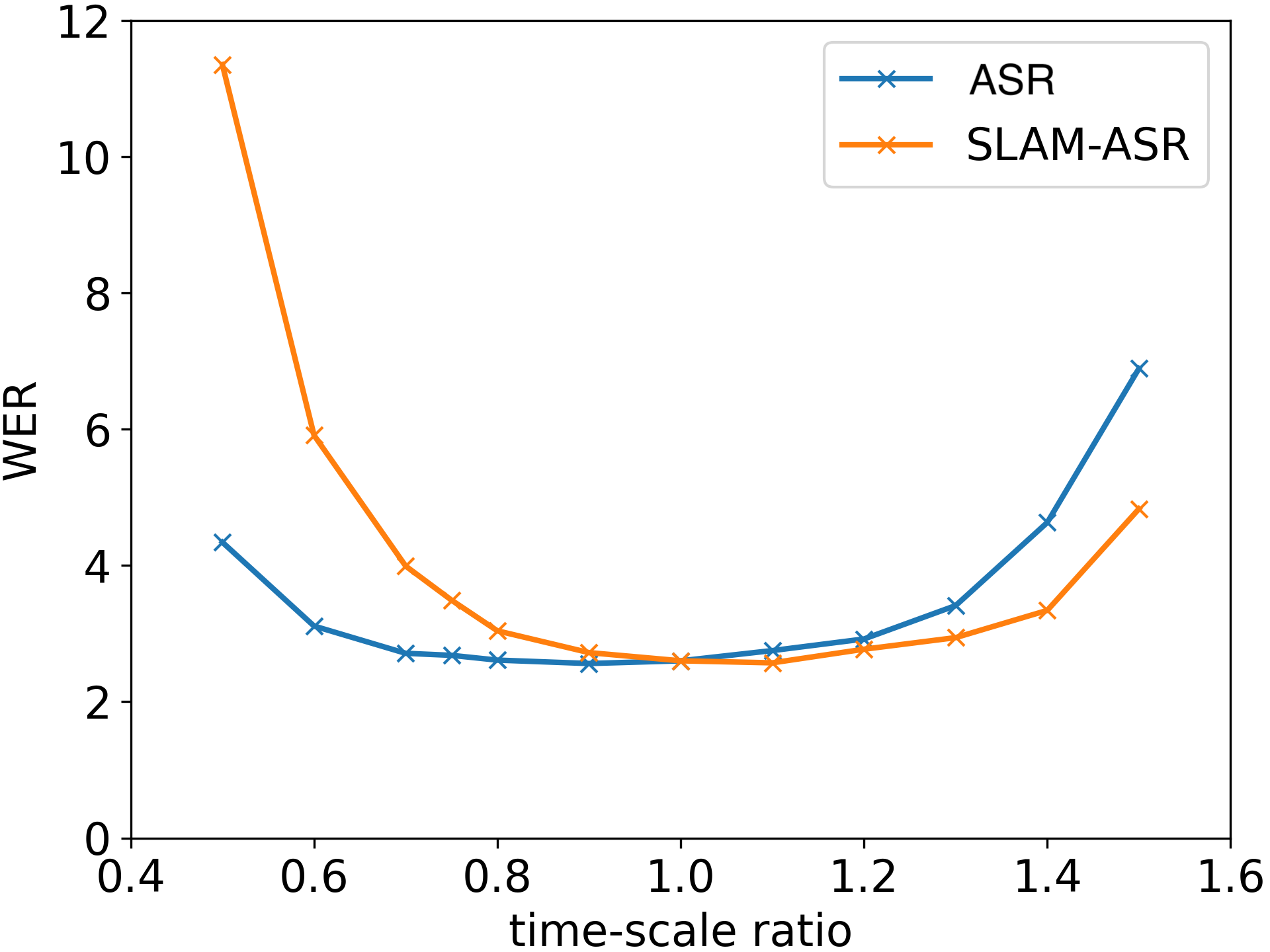}  
        \caption{\textit{Tempo}}
        \label{fig:tempo}
    \end{subfigure}
    \begin{subfigure}{.3\textwidth}
        \centering
        \includegraphics[width=.95\linewidth]{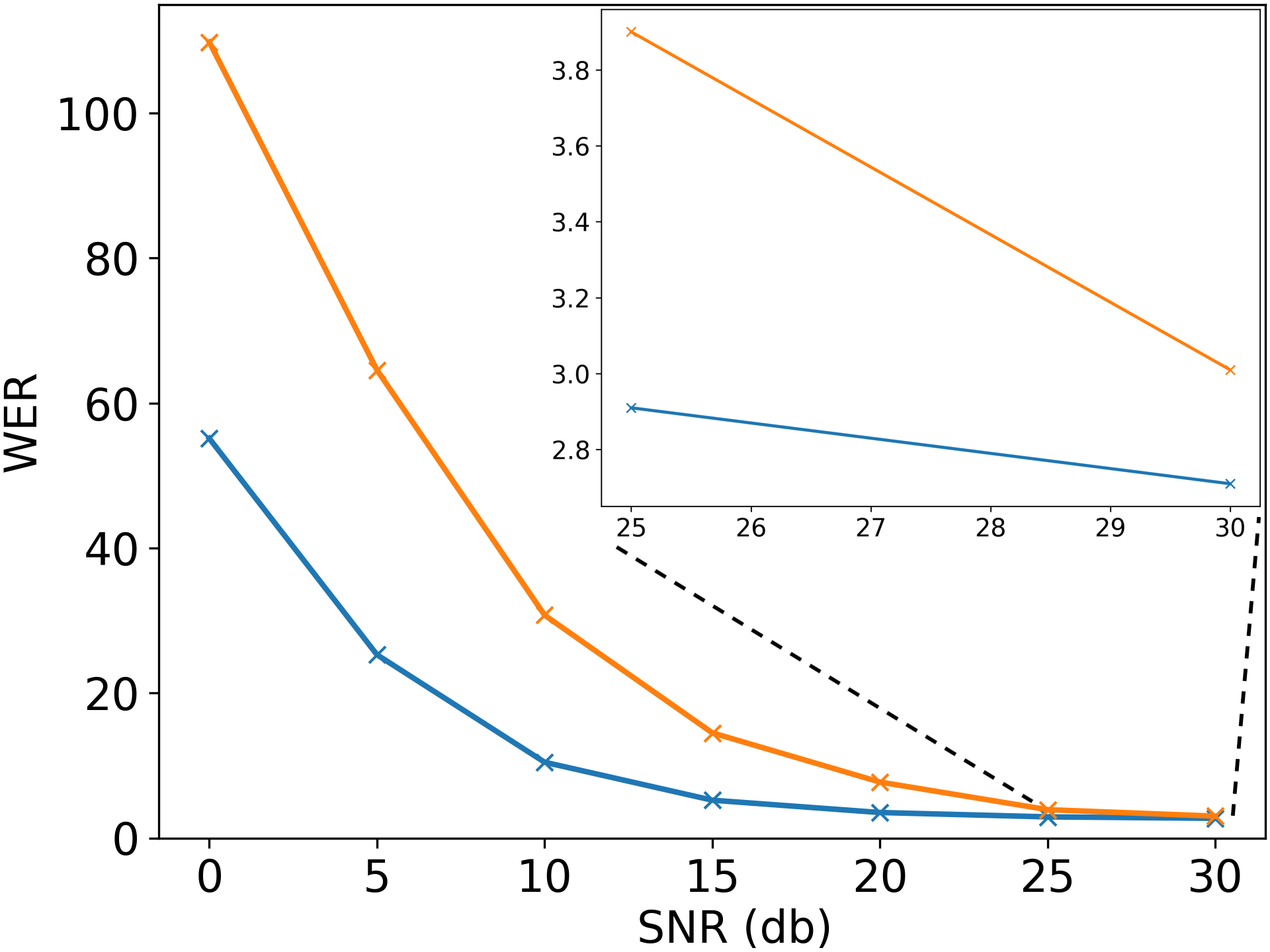}  
        \caption{\textit{Babble noise}}
        \label{fig:babble}
    \end{subfigure}
    \begin{subfigure}{.3\textwidth}
        \centering
        \includegraphics[width=.95\linewidth]{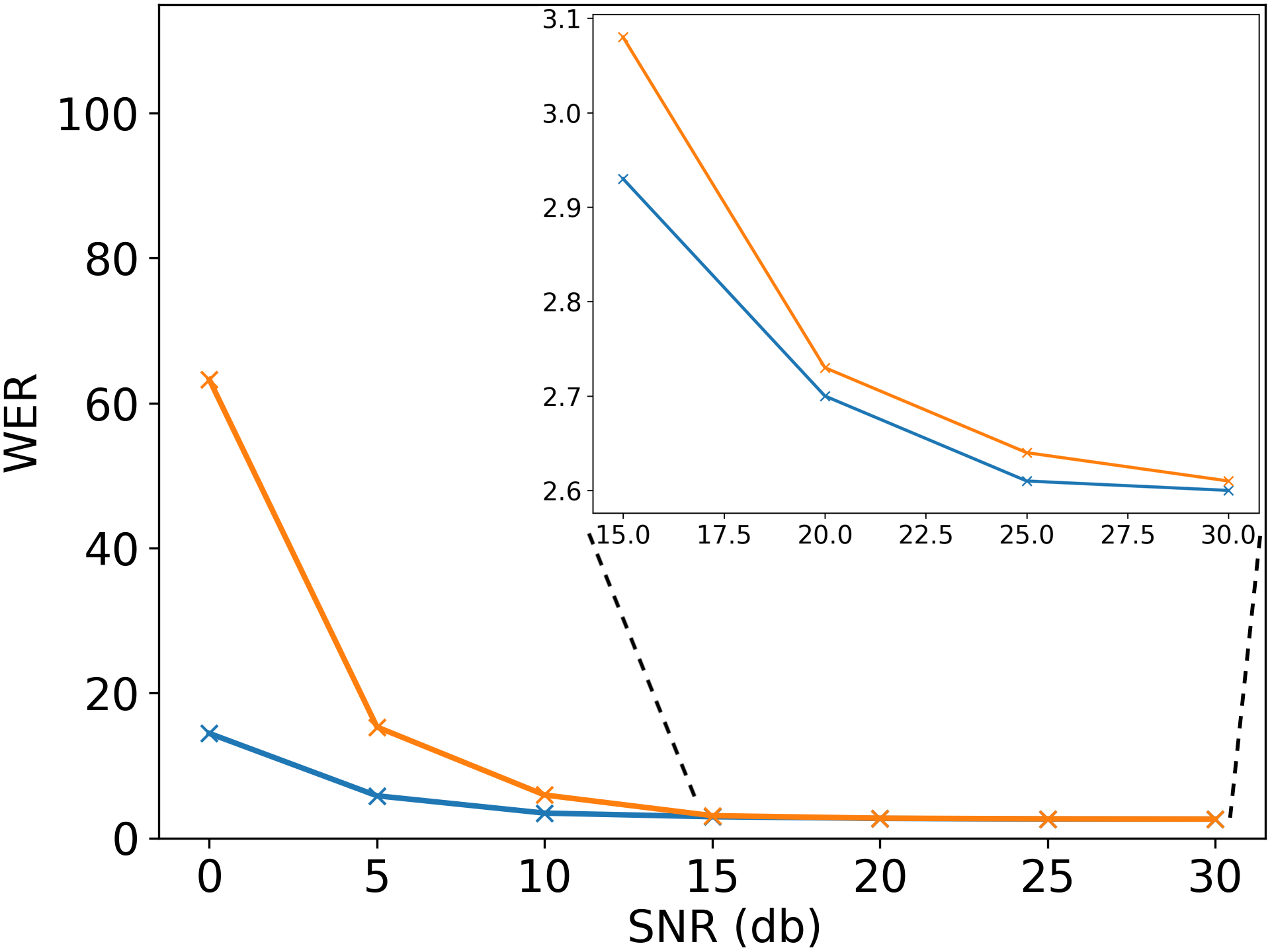}  
        \caption{\textit{Music noise}}
        \label{fig:music}
    \end{subfigure}
    \caption{Analysis of the impact of tempo (speech rate) and noise on WER.}
    \label{fig:wer-vs-tempo-vs-babble}
    \vspace{-4mm}
\end{figure*}

\begin{figure}[t]
    \centering
    \begin{subfigure}[b]{0.49\columnwidth}  
        \centering
        \includegraphics[width=\textwidth]{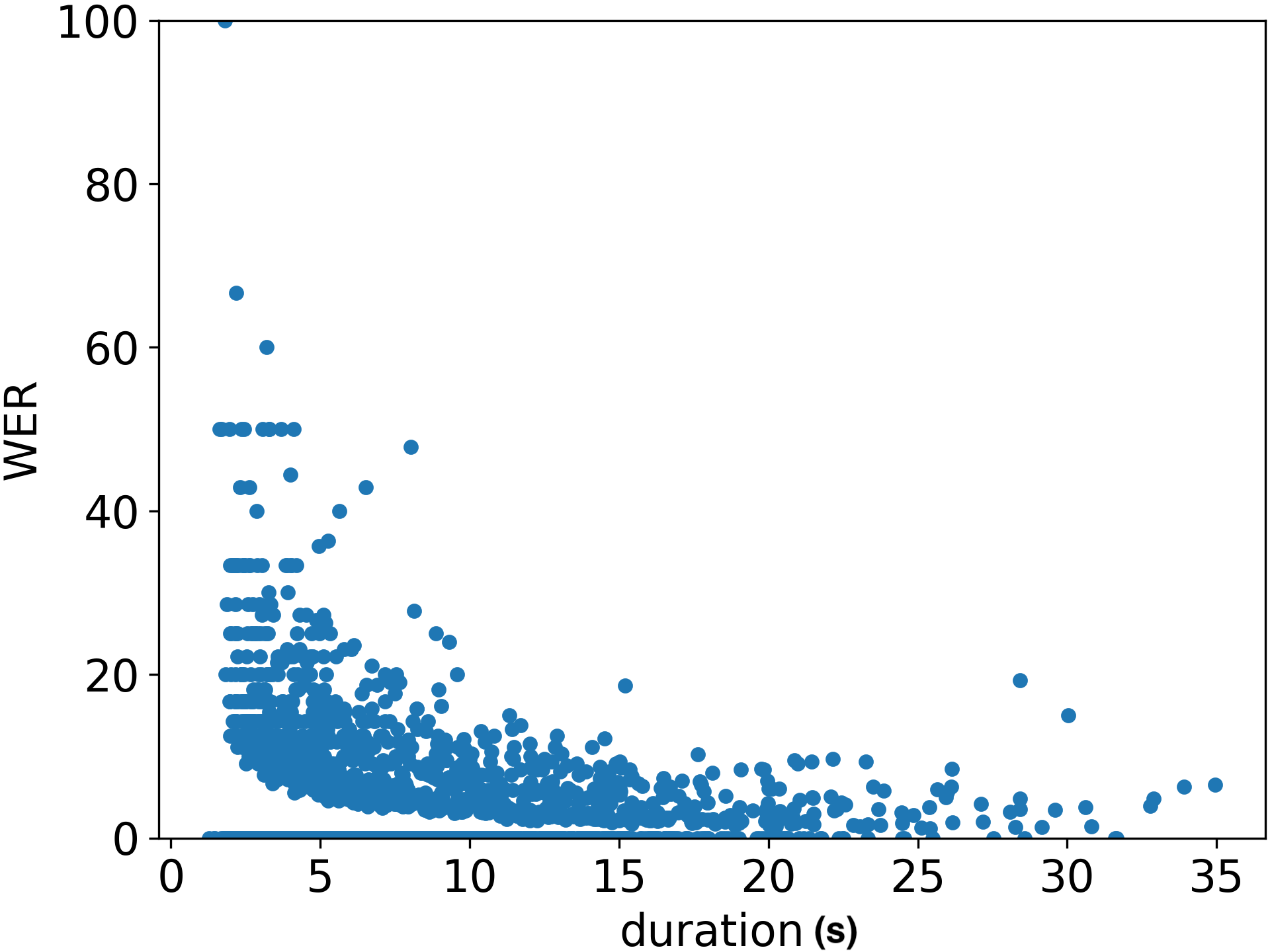}
        \caption{\textit{ASR (unchanged)}}
        \label{fig:subfig1}
    \end{subfigure}
    \begin{subfigure}[b]{0.49\columnwidth}
        \centering
        \includegraphics[width=\textwidth]{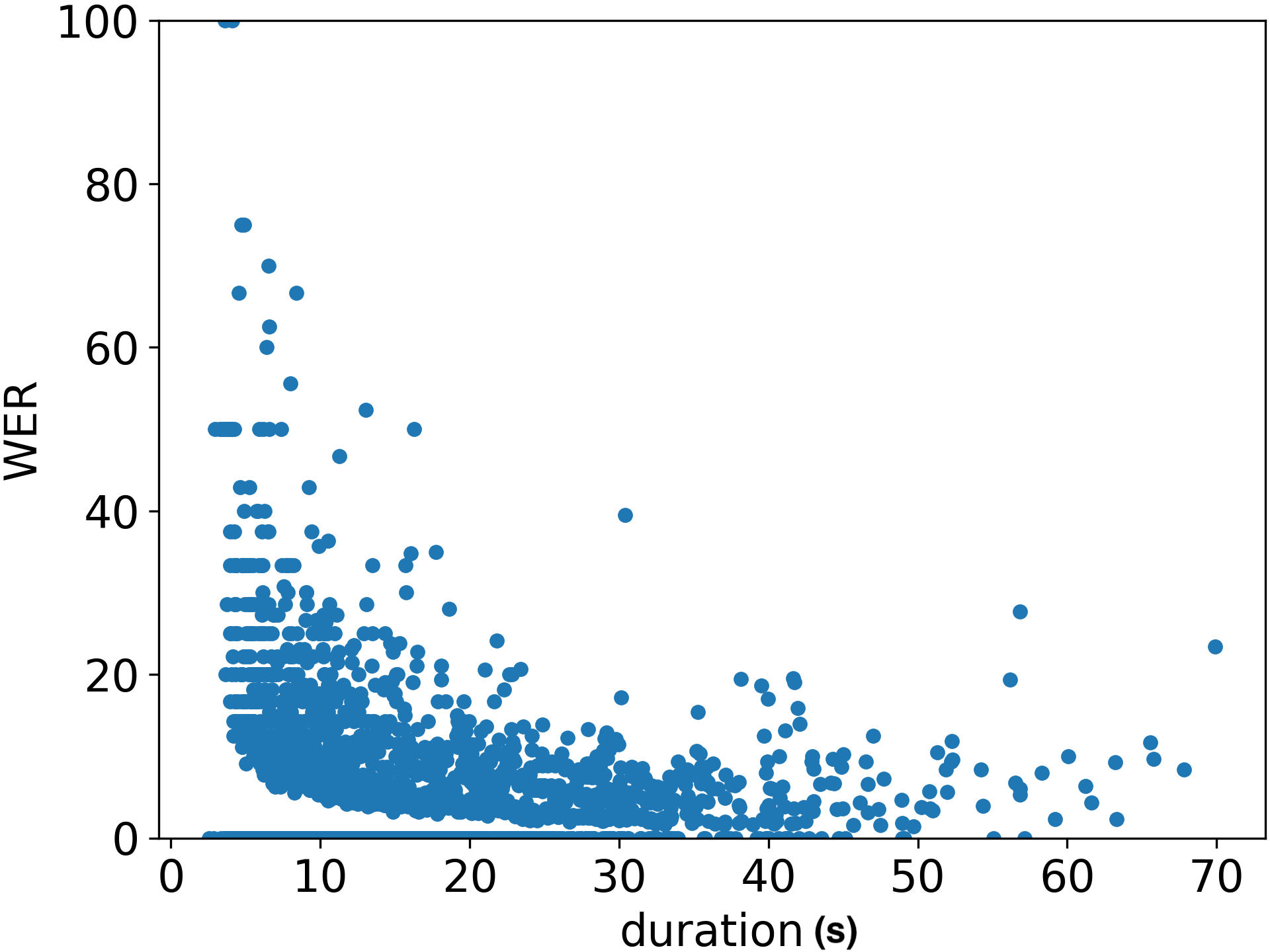}
        \caption{\textit{ASR (tempo=0.5)}}
        \label{fig:subfig2}
    \end{subfigure}
    
    
    \begin{subfigure}[b]{0.49\columnwidth}
        \centering
        \includegraphics[width=\textwidth]{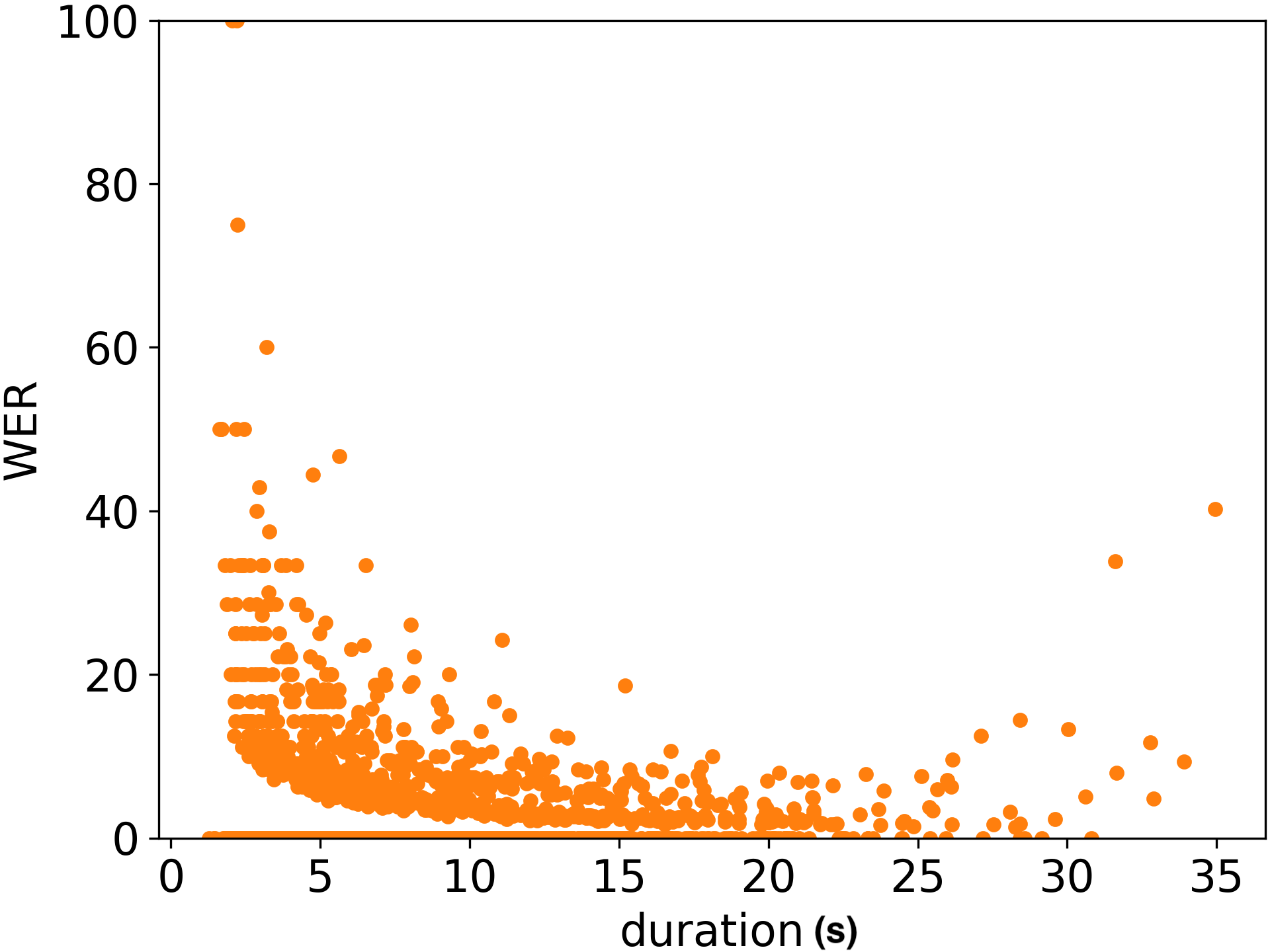}
        \caption{\textit{SLAM-ASR (unchanged)}}
        \label{fig:subfig3}
    \end{subfigure}
    \begin{subfigure}[b]{0.49\columnwidth}
        \centering
        \includegraphics[width=\textwidth]{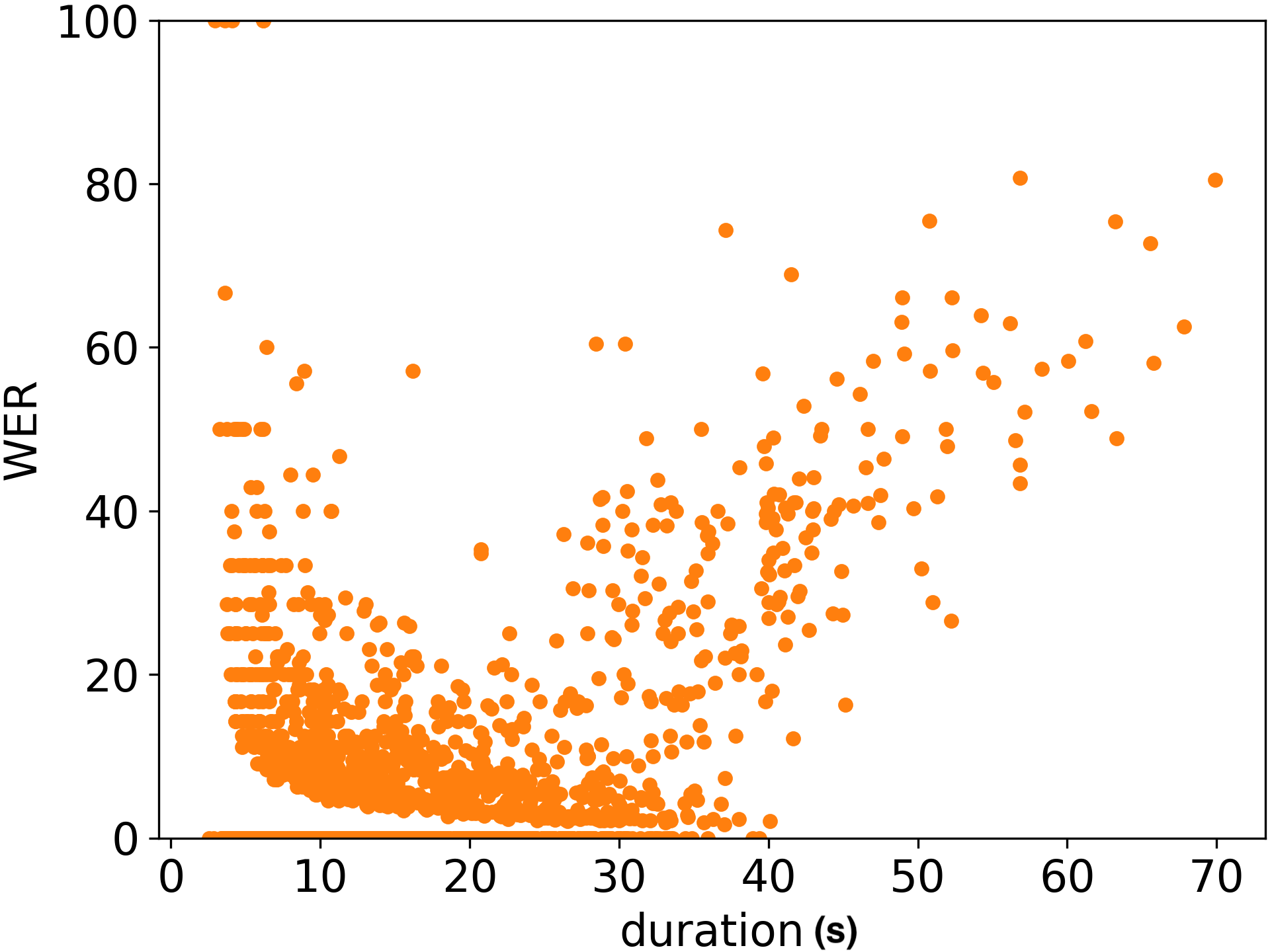}
        \caption{\textit{SLAM-ASR (tempo=0.5)}}
        \label{fig:subfig4}
    \end{subfigure}
    
    \caption{Scatter plots of WER versus speech duration for SLAM-ASR (bottom) and ASR baseline (top) on the LibriSpeech \textit{test-clean} set: unchanged (left) and half-speed (right).}
    \label{fig:correlation}
    \vspace{-4mm}
\end{figure}

\subsection{Training setup and technical details}
For the experiments on LibriSpeech, we use the publicly available checkpoint from SLAM-ASR\footnote{\url{https://github.com/X-LANCE/SLAM-LLM}}, while we trained the projectors for CallHome and \DefinedAI\ datasets in-house.
To ensure that our results are reproducible and comparable to SLAM-ASR results, in all our experiments, we followed the setup from the original work~\cite{ma2024embarrassingly}: the projector was trained for 3 epochs, with early stopping based on cross-entropy loss on the dev set, using AdamW~\cite{loshchilov2018decoupled} with a learning rate $\gamma=10^{-4}$ and a batch size of 4. The speech encoder produces output at 50 Hz and the downsampling rate is set to $k = 5$, leading to the downsampled audio features $\mathbf{Z}_i$ having a rate of 10 speech token embeddings $\mathbf{E}_i$ per second. The projector hidden layer dimension is set to 2048 and beam search with a beam size of 4 is used. 

\section{Experiments and Results}


\subsection{Cross-domain robustness evaluation}
\label{subsecc:cross-domain-results}

Since the projector is the only trainable component in SLAM-ASR, we hypothesize it may be prone to overfitting the domain seen during training.
This could limit its generalization to unseen domains or tasks when compared to standard ASR models, as highlighted in \cite{li2024whisma}.
Thus, the goal of this experiment is to evaluate the robustness of SLAM-ASR in cross-domain scenarios. 
More precisely, we assess the performance degradation of SLAM-ASR when trained on one dataset, but evaluated on different, mismatched datasets (Section~\ref{subsecc:dataset}).
We compare this degradation to that of standard ASR under the same conditions.



From the results, shown in Table \ref{tab:cross_domain_results}, it is clear that SLAM-ASR shows much greater performance degradation than its ASR counterpart when decoding audio from datasets different from those used in training, consistently across all datasets.
Additionally, SLAM-ASR exhibits exceptionally high WER (denoted by $\infty$) for certain dataset combinations.
Upon inspection, we identified that this behavior was mainly due to a substantial increase in insertion errors, caused by LLM hallucinations spiraling out of control.

\subsection{Speech perturbation ablations}


In the previous subsection, SLAM-ASR demonstrated limited cross-dataset generalization capabilities.
This suggests that the model's claimed strong speech recognition performance may be partially due to a reliance on dataset-specific acoustic and speaker characteristics, which it exploits as shortcuts to map audio features $\mathbf{Z}_i$ to speech token embeddings $\mathbf{E}_i$.
To further investigate this hypothesis, we designed a series of ablation experiments incorporating various speech perturbation techniques applied to the original evaluation audio.
Specifically, we evaluate how different levels of perturbations impact the performance of SLAM-ASR. 
We used the SLAM-ASR model trained on LibriSpeech, enabling a comparison of the results reported in the original SLAM-ASR paper~\cite{ma2024embarrassingly} with those obtained in this section, when the test waveforms are altered.

\subsubsection{Tempo perturbation} 


In SLAM-ASR, the projector must learn to map each audio feature $\mathbf{Z}_i$ to the LLM's input space, specifically the (sub)word embedding space.
We hypothesize that the learned mapping could be influenced by the speaking rates present in the training data.
Intuitively, the average speaking rate affects the average number of frames into which words are split and, consequently, the average number of audio feature $\mathbf{Z}$s to be mapped to the words' corresponding $\mathbf{E}$s.
Therefore, we assess the robustness of SLAM-ASR to time-scale modifications of the test audio.
In particular, the time-scale was modified by simply manipulating the hop length using the well-established PSOLA method~\cite{psola,parselmouth} which modifies the prosody of natural speech while retaining a high level of naturalness.

Figure~\ref{fig:tempo} shows the change in WER performance across time-scale ratios ranging from 0.5 to 1.5, in increments of 0.1.
As the speaking rate increases (\textit{\textit{i.e.}}, $ratio > 1$), both models exhibit a similar decline in performance. However, when the rate slows down (\textit{\textit{i.e.}}, $ratio < 1$), SLAM-ASR’s WER rises sharply, reaching nearly 12\% at a ratio of 0.5 (half the normal rate), in contrast to its ASR counterpart. These results suggest that SLAM-ASR struggles to handle an increase in the number of frames that must be mapped to the same number of (sub)words. This phenomenon may occur because a greater mismatch between the number of frames and the LLM's input wordpieces makes the mapping from $\mathbf{Z}$ to the corresponding $\mathbf{E}$ significantly more challenging.
To further support this hypothesis, in Figure~\ref{fig:correlation} we plot utterance-level WER values against speech duration.
We can see that SLAM-ASR's WER starts to increase linearly when audio inputs are longer than $\approx30$ seconds (\textit{\textit{i.e.}} longer than 300 $\mathbf{Z}$s audio features), unlike ASR, which maintains a consistent WER pattern regardless of rate changes\footnote{Note that this behavior is unlikely due to reaching the maximum LLM context-length, 4K for Vicuna-7B, since the longer audio samples generate only around 700 speech tokens.}

\subsubsection{Noise augmentation}

Given that the LLM and speech encoder are fixed in SLAM-ASR, the projector must be robust to relatively small changes in audio feature $\mathbf{Z}_i$ for accurate mapping to the LLM's embedding space.
We hypothesize that the projector’s tendency to find shortcuts for this mapping may significantly impact SLAM-ASR’s performance, even with simple additive noise in the audio.
Therefore, we add noise to the waveforms\cite{noise-augmentation} to assess SLAM-ASR's robustness to noise.
We use music and speech (babble) noise files from the MUSAN \cite{snyder2015musan} corpus.

Figures~\ref{fig:babble} and \ref{fig:music} show the WER when babble and music noise,
respectively, are added,  resulting in a Signal-to-Noise Ratio (SNR) ranging from 0 to 30 decibels (dB) in 5 dB increments.
As the noise increases (30dB $\rightarrow$ 0dB), the performance of SLAM-ASR degrades quite rapidly in contrast to its ASR counterpart.
With babble noise in particular, even a small amount of noise (\textit{e.g.}, 20 dB) causes substantial degradation in SLAM-ASR's performance, while the ASR model experiences only minimal degradation.
These results suggest SLAM-ASR struggles with minor audio changes unseen in training.

\subsection{Speech-to-text alignment analysis}



\begin{figure}[t!]
    \centering
    \includegraphics[width=\linewidth]{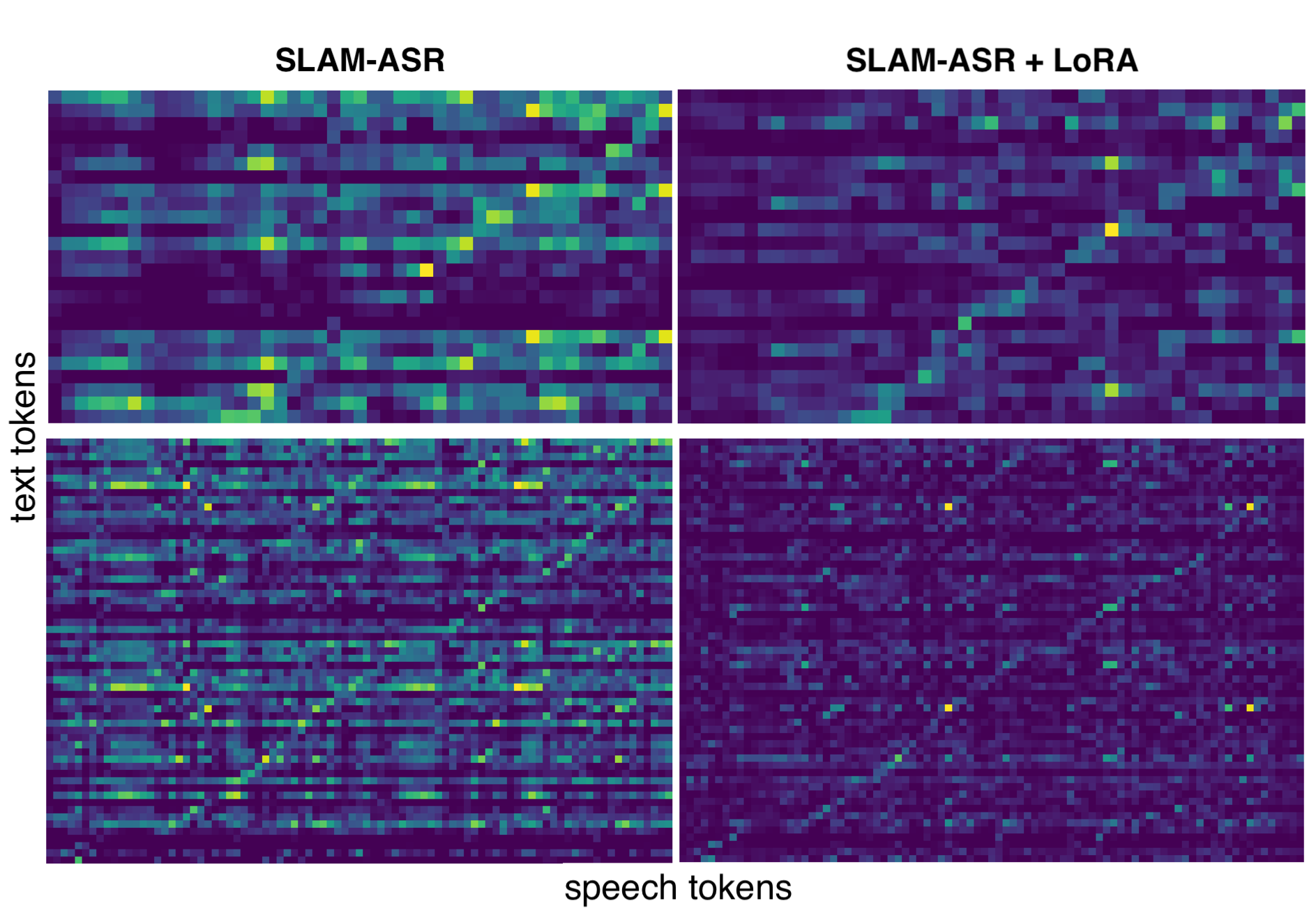}
    \caption{The pairwise cosine similarity between every pair of speech and text token embeddings for two test examples before (left side) and after using LoRA (right side) in the LLM. Colors range from purple (-1, low similarity) through green (0, neutral) to yellow (+1, high similarity), using the viridis colormap.}
    \label{fig:alignment}
    \vspace{-2mm}
\end{figure}
\begin{figure}[t!]
    \centering
    \includegraphics[width=\linewidth]{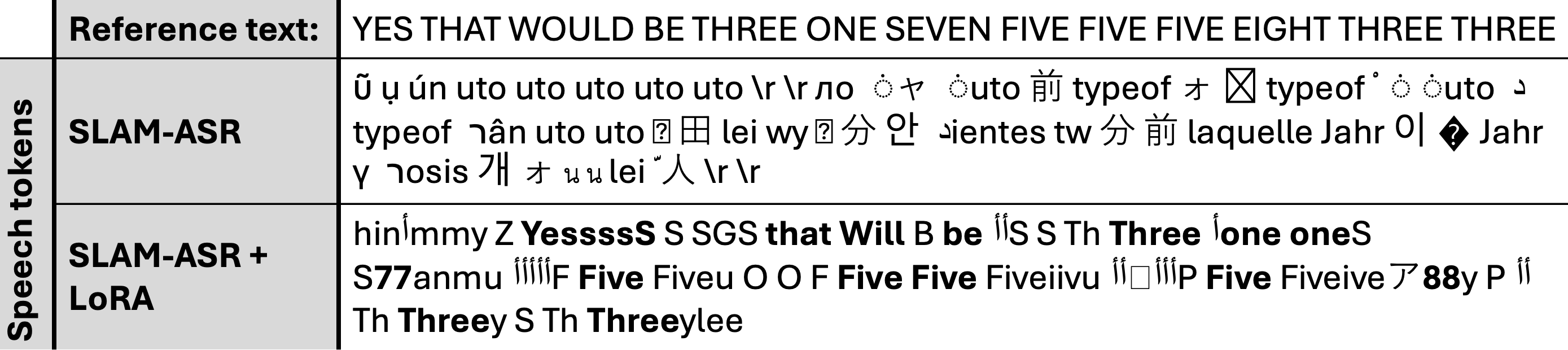}
    \caption{Ground truth transcript compared to learned speech tokens for SLAM-ASR and SLAM-ASR+LoRA for the same input audio.}
    \label{fig:aligned_tokens_example}
    \vspace{-4mm}
\end{figure}

Previous research has suggested that the LLM-based ASR task can be viewed as a regurgitation activity, where the language model is responsible for refining and reproducing information in the same order as it appears in the audio encoder's output sequence~\cite{fathullah2024prompting}. Thus, if the projector in SLAM-ASR can provide a sequence of embeddings monotonically aligned with the text embeddings, the LLM-based ASR problem reduces to a repetition task, which should not require the full capacity of an LLM. Nevertheless, a key aspect of the SLAM-ASR architecture is that both the speech encoder and the LLM are frozen, which contrasts with the current trend in LLM-based ASR research~\cite{fathullah2024prompting, das2024speechverse, tangsalmonn, wu2023decoder, ma2024embarrassingly}, where adapters (\textit{e.g.}, LoRA~\cite{hu2022lora}) are used to fine-tune the LLM. Hence, we hypothesize that SLAM-ASR will face more difficulties in learning the alignment of speech tokens and text tokens compared to methods that also fine-tune the LLM. To validate our hypothesis, we conducted an additional experiment on the \DefinedAI\ dataset using the LoRA adapter to fine-tune the LLM alongside training the projector of SLAM-ASR. We then computed the cosine similarity between each possible pair of speech tokens and text tokens embedding of the LLM. Figure \ref{fig:alignment} shows the resultant alignment plots for two randomly selected samples. As can be inferred from the similarity plots, the task of aligning audio to text is harder for the original SLAM-ASR setup (left-side plots).

Additionally, to assess what the projector is learning, we mapped the learned speech tokens to their closest token from the LLM's vocabulary. One example from this exploration is shown in Figure \ref{fig:aligned_tokens_example}. Note that the output from SLAM-ASR is mostly gibberish, while the tokens retrieved by SLAM-ASR+LoRA are much more aligned with the reference text. Overall, the results from this ablation raise the question whether freezing the LLM is advisable, given the clear advantage of using LoRA for better alignments and the improvement in performance for in-domain and cross-domain scenarios.\footnote{SLAM-ASR+LoRA experiments using \DefinedAI\ as training and evaluation data (in-domain setup) improved the WER from 13.8 to 11.6. For the cross-domain setup, \textit{i.e.},  (\textit{train}) \DefinedAI\ $\rightarrow$ (\textit{test}) LibriSpeech, WER improves from 60.4 to 22.5. We had to switch the default fp16 to fp32 precision for training with LoRA, as the default setting did not converge.}

\section{Conclusions}
\label{sec:conclusions}
We have investigated what is and what is not advisable when working with a recent, widely adopted LLM-based ASR solution, \textit{i.e.}, SLAM-ASR. By ablation, we were able to identify the good, the bad, and the ugly aspects of this ASR paradigm in which both the speech foundation model and the LLM are frozen, and connected through a trainable projector. 

\textbf{The Good:} SLAM-ASR is efficient, both in computation and data usage, while delivering competitive in-domain performance. Its flexibility allows easy integration of various speech encoders and LLMs, the only trainable component being a simple projector, a great advantage when computational resources or data are scarce. 

\textbf{The Bad:} Like many other LLM-based ASR systems, SLAM-ASR tends to strongly overfit the training domain. This is a clear disadvantage compared to traditional ASR models, which have better generalization capabilities and are less sensitive to domain shifts and speech perturbation. Our experiments show that SLAM-ASR can easily go off the rails and hallucinate words (\textit{e.g.}, repetitions or unrelated text) when simple perturbations are applied to the speech signal. 

\textbf{The Ugly:} Unlike LLM-based ASR approaches that use LoRA for fine-tuning the LLM, the SLAM-ASR architecture lacks clear evidence that the projector is learning an alignment between speech and text, rather than some other spurious correlation. Our analysis shows that the alignment is less evident without LoRA, resulting in gibberish output when speech tokens are mapped to text tokens. 

\textbf{The Way Forward:} Overall, our ablations suggest that SLAM-ASR should be trained and used with in-domain data for inference. However, if the data is very noisy, such as in CallHome, a traditional ASR model is the better choice. Similarly, our experimental analysis suggests that LoRA adapters should be considered for improving the alignment of speech and text tokens, as well as for better performance of SLAM-ASR in both in-domain and cross-domain scenarios. We recognize that many factors can influence the learning capabilities of LLM-based ASR models; however, we believe that our insights will benefit the research community working in similar areas. 

\section*{Acknowledgments}
This work was supported by the Idiap~\&~Uniphore collaboration project.
Part of the work was also supported by EU Horizon 2020 project ELOQUENCE\footnote{\url{https://eloquenceai.eu/}} (grant number 101070558).

\balance
\bibliographystyle{IEEEtran}
\bibliography{references}

\end{document}